\documentclass[letterpaper]{article}
\usepackage[draft]{aaai2026}
\usepackage{times} 
\usepackage{helvet}
\usepackage{courier}
\usepackage[hyphens]{url}
\usepackage{graphicx}

\usepackage{amsfonts,amssymb}
\usepackage{multirow}
\usepackage{natbib}
\usepackage{caption}
\usepackage{algorithm}
\usepackage{algorithmic}
\usepackage{amsmath}
\usepackage{booktabs}

\usepackage{newfloat}
\usepackage{listings}
\DeclareCaptionStyle{ruled}{labelfont=normalfont,labelsep=colon,strut=off}
\floatstyle{ruled}
\newfloat{listing}{tb}{lst}{}
\floatname{listing}{Listing}

\title{GS: Generative Segmentation via Label Diffusion}
\author{
    Yuhao Chen \textsuperscript{\rm 1}, Shubin Chen \textsuperscript{\rm 1},  Liang Lin \textsuperscript{\rm 1} \textsuperscript{\rm 2} \textsuperscript{\rm 3}, Guangrun Wang \textsuperscript{\rm 1} \textsuperscript{\rm 2} \textsuperscript{\rm 3}\footnotemark[2]
    \\
}
\affiliations{
    \textsuperscript{\rm 1}Sun Yat-sen University,
    \textsuperscript{\rm 2} Guangdong Key Laboratory of Big Data Analysis and Processing, \textsuperscript{\rm 3} X-Era AI Lab,
}

\usepackage{bibentry}

\begin{document}

\maketitle
\renewcommand{\thefootnote}{\fnsymbol{footnote}} 
\footnotetext[2]{Corresponding author.}

\begin{abstract}
Language-driven image segmentation is a fundamental task in vision-language understanding, requiring models to segment regions of an image corresponding to natural language expressions. Traditional methods approach this as a discriminative problem, assigning each pixel to foreground or background based on semantic alignment. Recently, diffusion models have been introduced to this domain, but existing approaches remain image-centric: they either (i) use image diffusion models as visual feature extractors, (ii) synthesize segmentation data via image generation to train discriminative models, or (iii) perform diffusion inversion to extract attention cues from pre-trained image diffusion models—thereby treating segmentation as an auxiliary process.
In this paper, we propose \textbf{GS (Generative Segmentation)}, a novel framework that formulates segmentation itself as a generative task via \textit{label diffusion}. Instead of generating images conditioned on label maps and text, GS reverses the generative process: it directly generates segmentation masks from noise, conditioned on both the input image and the accompanying language description. This paradigm makes label generation the primary modeling target, enabling end-to-end training with explicit control over spatial and semantic fidelity.
To demonstrate the effectiveness of our approach, we evaluate GS on \textit{Panoptic Narrative Grounding (PNG)}, a representative and challenging benchmark for multimodal segmentation that requires panoptic-level reasoning guided by narrative captions. Experimental results show that GS significantly outperforms existing discriminative and diffusion-based methods, setting a new state-of-the-art for language-driven segmentation. 

\end{abstract}

\begin{figure}[t]
\vspace{-11pt}
    \centering
    \includegraphics[width=0.9\linewidth]{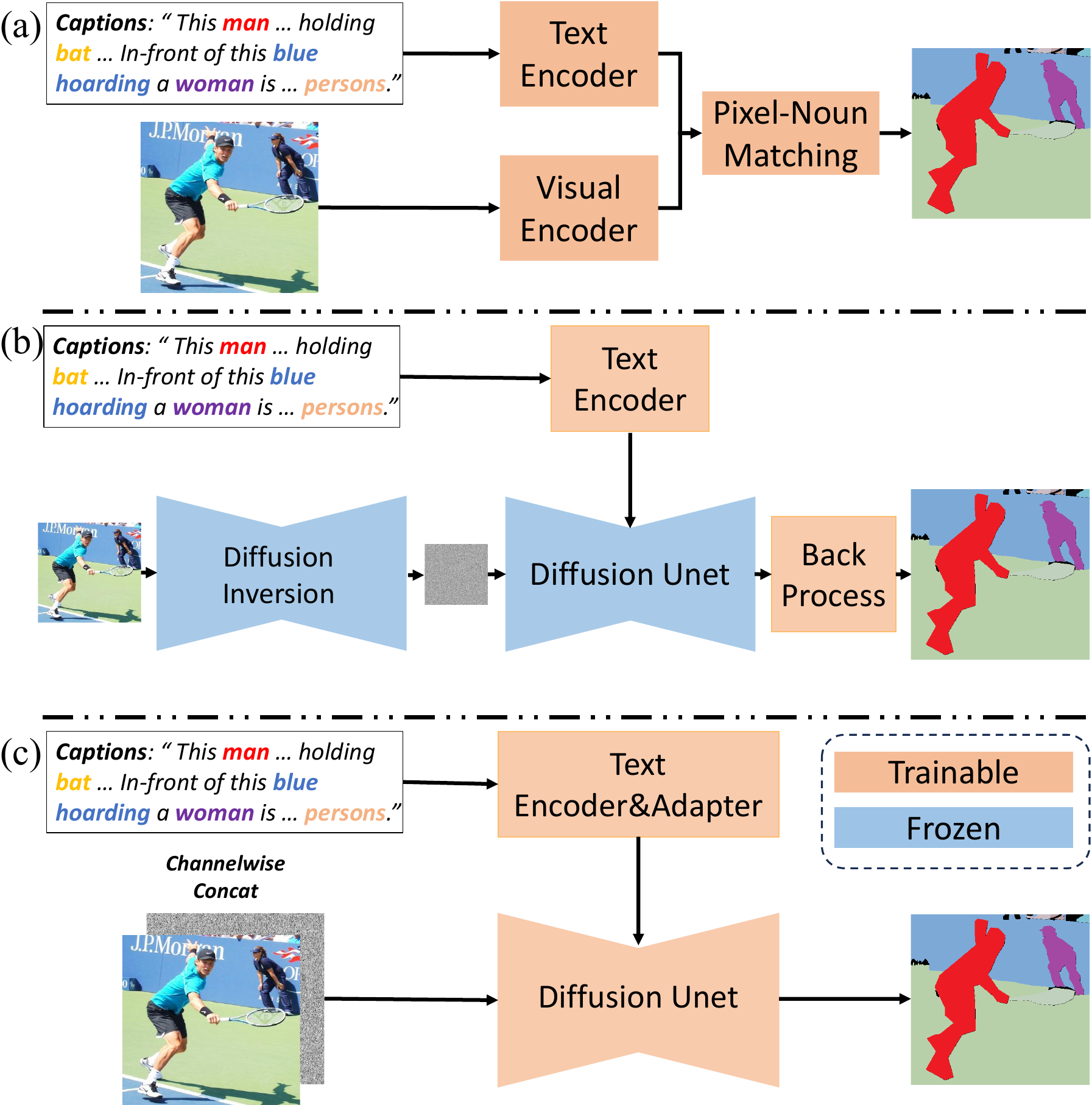}
    \caption{\small{\textbf{Comparison of GS with prior segmentation paradigms.} 
(a) Discriminative models encode image and text into feature space and predict pixel-wise segmentation; this includes both traditional methods and diffusion-assisted approaches that rely on external discriminators. 
(b) Diffusion inversion methods extract attention maps from pre-trained image diffusion models for zero-shot grounding. 
(c) Our proposed GS directly performs label-space diffusion to generate segmentation masks from noise, conditioned on image and text, enabling end-to-end training without auxiliary components.
}}
    \label{fig:insight}
    \vspace{-20pt}
\end{figure}

\section{Introduction}
\label{sec:intro}

Image segmentation conditioned on natural language descriptions—often referred to as language-driven segmentation—has emerged as a critical capability for multimodal intelligence. Unlike traditional segmentation tasks that rely solely on visual cues, this task demands understanding and aligning both spatial and semantic information across modalities. Applications span referring expression segmentation, open-vocabulary mask generation, and panoptic narrative grounding.

Recent advances in diffusion models \cite{ho2020denoising,nichol2021improved} have led to remarkable progress in generative modeling. Models like Stable Diffusion \cite{rombach2022high} and ControlNet \cite{zhang2023adding} have demonstrated strong spatial reasoning and flexible conditioning capabilities, motivating their application to vision-language tasks. However, most diffusion-based segmentation methods remain fundamentally tied to the \textit{image generation} process. Specifically, current approaches either: (1) extract features from frozen diffusion backbones for downstream discriminative models \cite{zhu2024exploring}, (2) generate synthetic segmentation data via diffusion to supervise segmentation models \cite{li2023open,nguyen2023dataset,ma2023diffusionseg,wu2023diffumask}, or (3) perform inversion of image diffusion to extract intermediate attention maps for zero-shot grounding \cite{yang2024exploring,liu2024vgdiffzero,ni2023ref}. These designs ultimately treat segmentation as an auxiliary or post-hoc process, not the generative objective itself.

In this paper, we propose a new paradigm—\textbf{Generative Segmentation (GS)}—which reframes language-driven segmentation as a primary generative task. Rather than using diffusion to generate images, \textbf{GS} directly learns to generate segmentation masks from noise through a label-space diffusion process. Conditioned on both the input image and the language description, GS predicts a denoised segmentation map, effectively grounding linguistic concepts into precise pixel-wise masks. This task reversal—termed \textit{label diffusion}—enables GS to perform structured mask generation directly, trained end-to-end without auxiliary stages or handcrafted post-processing.

To evaluate our method, we adopt \textit{Panoptic Narrative Grounding (PNG)} as a representative benchmark. PNG is one of the most comprehensive language-driven segmentation tasks to date, requiring models to segment both ``things'' and ``stuff'' in an image based on noun phrases extracted from free-form narrative captions.

A comparison between conventional panoptic narrative grounding methods and our label diffusion approach is illustrated in Fig.~\ref{fig:insight}.

\noindent\textbf{Our contributions are as follows:}
\begin{enumerate}
    \item We propose \textbf{GS}, a novel generative segmentation framework based on \textit{label diffusion}, which directly generates segmentation masks conditioned on image and text inputs.
    \item We introduce a dual-branch conditioning mechanism that injects spatial structure and semantic context into the generative process, enabling precise and faithful segmentation.
    \item We demonstrate that GS achieves state-of-the-art performance on the Panoptic Narrative Grounding (PNG) benchmark, validating the strength and generality of our formulation.
\end{enumerate}

\section{Related Work}

\paragraph{Diffusion Models for Segmentation.}
Diffusion models~\cite{ho2020denoising,sohl2015deep} have become the foundation of high-fidelity image generation, achieving unprecedented performance in tasks such as text-to-image synthesis~\cite{rombach2022high} and structured conditioning~\cite{zhang2023adding}. Their capacity for spatial alignment and multimodal reasoning has inspired research into adapting them for vision-language segmentation. Existing efforts typically incorporate diffusion models in one of three ways: (1) as frozen feature extractors for downstream discriminative models~\cite{zhu2024exploring}, (2) as generative engines for producing synthetic data to supervise segmentation models~\cite{li2023open,nguyen2023dataset,ma2023diffusionseg,wu2023diffumask}, or (3) through inversion techniques that extract attention maps for zero-shot localization~\cite{yang2024exploring,liu2024vgdiffzero,ni2023ref}. While effective, these designs use image generation as an intermediary and treat segmentation as a derived or auxiliary target.

In contrast, our work introduces \textbf{label diffusion}, which treats the segmentation mask itself as the object of generation. Rather than leveraging diffusion indirectly, our proposed GS model directly denoises segmentation masks from noise, conditioned on both visual and linguistic inputs. This formulation positions segmentation as a primary generative task, enabling explicit, structured alignment between image regions and language without reliance on handcrafted guidance or pretraining on external generation objectives. To our knowledge, GS is the first framework to perform language-driven segmentation via direct mask synthesis using conditional diffusion.

\paragraph{Generative Models for Visual Understanding.}
The idea of modeling data distributions to improve discriminative performance has a long history in computer vision. Early works~\cite{hinton2007recognize,ng2001discriminative} demonstrated that learning generative representations, could enhance downstream tasks like image classification by uncovering latent structure in the data. More recently, generative models have proven effective for learning representations applicable to both global and dense prediction tasks, including classification~\cite{li2023mage,wang2022semantic} and segmentation~\cite{baranchuk2021label}.
However, most of these approaches either train discriminative and generative objectives jointly, or fine-tune generative backbones for downstream tasks—treating generation as a pretext task rather than a means of direct prediction.

Diffusion models~\cite{ho2020denoising,sohl2015deep} have emerged as particularly expressive generative frameworks, achieving strong results in image synthesis~\cite{dhariwal2021diffusion,rombach2022high}, while exhibiting powerful spatial reasoning and cross-modal alignment. Recent extensions such as ControlNet~\cite{zhang2023adding} introduce controllable generation using structural priors like edges and depth, while pre-trained models such as Segment Anything~\cite{kirillov2023segment} leverage foundation vision models for open-vocabulary region parsing. Nevertheless, the majority of prior work leverages diffusion models indirectly—either as feature extractors, data generators, or through inversion pipelines—and does not directly use the generative process itself to perform task-specific predictions like segmentation or classification \cite{li2023your}.

In this work, we take a different route. Instead of treating generation as auxiliary, we model segmentation as a primary generative objective via a conditional label-space diffusion process. Our method, GS, demonstrates that diffusion models can directly synthesize structured outputs such as segmentation masks from noise, conditioned on image and text. This contributes to a relatively underexplored direction: using generative models not just for representation learning, but for direct, interpretable predictions.

\begin{figure*}[t]
    \centering
    \includegraphics[width=0.9\linewidth]{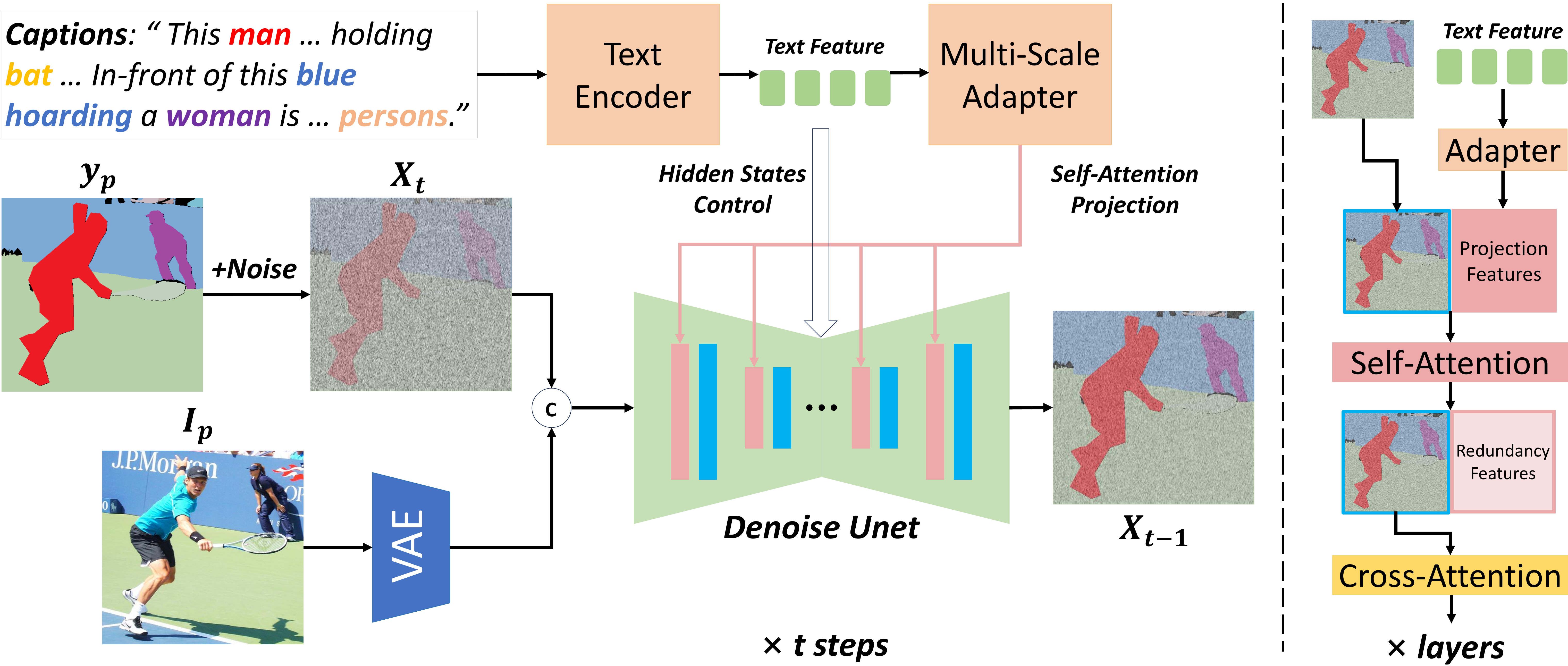}
    \caption{\textbf{Overview of GS.} Our Generative Segmentation (GS) framework synthesizes segmentation masks via a conditional label-space diffusion process. During training, ground-truth masks are noised in the latent space and denoised by a U-Net conditioned on both the input image and the associated text. The image is encoded by a VAE and concatenated channel-wise to the noisy label. Text guidance is incorporated through global CLIP embeddings and multi-scale adapter features, which are injected into the U-Net via token-level attention as well as cross-attention. We apply classifier-free guidance by randomly dropping text conditions during training and combining conditional and unconditional predictions at test time. GS directly generates high-quality segmentation masks without relying on intermediate RGB synthesis.}
    \label{fig:overview}
    \vspace{-11pt}
\end{figure*}

\paragraph{Panoptic Narrative Grounding.}
Panoptic Narrative Grounding (PNG)~\cite{gonzalez2021panoptic} is a challenging vision-language task that generalizes referring segmentation to panoptic-level coverage. It requires associating free-form narrative captions with precise segmentation masks of both ``thing'' objects and ``stuff'' regions. Compared to traditional referring expression segmentation, PNG incorporates multiple noun phrases per image and demands modeling of complex linguistic structures including coreference, spatial relations, and attribute composition. Prior approaches primarily adopt discriminative pipelines that combine cross-modal encoders with per-phrase segmentation heads~\cite{gonzalez2021panoptic,li2024dynamic}. Recent variants enhance grounding quality using vision-language pretraining (e.g., CLIP) or use image generation diffusion models for feature encoding. However, these designs do not model mask synthesis as a generative process.

We adopt PNG as a representative benchmark to evaluate our proposed GS framework due to its comprehensive multimodal nature. The complexity of PNG highlights GS’s capability in simultaneously resolving linguistic reference and spatial mask generation through a unified, end-to-end generative process.

\section{Methods}

We introduce \textbf{GS (Generative Segmentation)}, a framework that directly generates segmentation masks via a conditional label-space diffusion process. Unlike previous approaches that rely on image synthesis pipelines, GS treats segmentation as a primary generative task. At the core of GS is a denoising model that generates pixel-wise masks from noise, guided jointly by the input image and a textual description. Fig.~\ref{fig:overview} provides an overview of our framework.

We evaluate GS on the Panoptic Narrative Grounding (PNG) task, a comprehensive benchmark involving language-conditioned segmentation. However, GS is general and applicable to a wide range of language-driven segmentation settings.

\subsection{Notation}
\label{sec:notation}

We summarize the key notations used throughout this paper in Table~\ref{tab:notation}. These notations define the image, language, and diffusion components used in our generative segmentation framework.

\begin{table}[h]
\centering
\resizebox{.48\textwidth}{!}{
\begin{tabular}{cl}
\toprule
\textbf{Symbol} & \textbf{Description} \\
\midrule
\( \mathcal{I} \in \mathbb{R}^{H \times W \times 3} \) & Input image \\
\( \mathcal{T} \) & Natural language caption associated with \( \mathcal{I} \) \\
\( T_i \) & The \( i \)-th noun phrase extracted from \( \mathcal{T} \) \\
\( y_i \in \mathbb{R}^{H \times W} \) & Ground-truth segmentation mask for \( T_i \) \\
\( \tilde{y}_i \in \mathbb{R}^{H \times W} \) & Predicted segmentation mask for \( T_i \) \\
\( \varepsilon(\cdot) \) & Variational autoencoder used for latent encoding \\
\( x_0 \) & Clean label (segmentation mask) in latent space \\
\( x_t \) & Noisy label at diffusion timestep \( t \) \\
\( \epsilon \sim \mathcal{N}(0, I) \) & Standard Gaussian noise \\
\( \bar{\alpha}_t \) & Predefined noise schedule at timestep \( t \) \\
\( e^t_i \) & Text embedding of noun phrase \( T_i \) from CLIP encoder \\
\( e^p_{il} \) & Layer-wise projected feature from adapter at U-Net layer \( l \) \\
\( \Phi(\cdot) \) & U-Net denoising network \\
\( X_l \) & Input feature at layer \( l \) of U-Net \\
\bottomrule
\end{tabular}
}
\caption{\textbf{Summary of notations used in GS.}}
\label{tab:notation}
\vspace{-12pt}
\end{table}

\subsection{Preliminaries}

\paragraph{Language-Driven Segmentation.}
Language-driven segmentation refers to the task of generating pixel-wise segmentation masks \( y_i \in \mathbb{R}^{H \times W} \) that correspond to natural language expressions \( T_i \), where \( i \) denotes the index of the \( i \)-th noun phrase in a given textual description. Given an input image \( \mathcal{I} \in \mathbb{R}^{H \times W \times 3} \) and an associated caption \( \mathcal{T} \), the objective is to identify and localize all referential noun phrases \( T_i \) within \( \mathcal{T} \), and to produce accurate segmentation masks \( y_i \) aligned with each phrase. In this work, we adopt the Panoptic Narrative Grounding (PNG) benchmark~\cite{gonzalez2021panoptic} as the primary evaluation setting, as it provides dense linguistic supervision and panoptic-level mask annotations.

\paragraph{Diffusion Modeling.}
Diffusion models~\cite{ho2020denoising} are a class of generative models that learn to approximate complex data distributions through a denoising process. They consist of two stages: a forward (noising) process that gradually adds Gaussian noise to the input data over a series of timesteps, and a reverse (denoising) process learned by a neural network. Given a clean data sample \( x_0 \), the forward process generates a noisy version \( x_t \) at timestep \( t \) as:
\begin{equation}
    x_t = \sqrt{\bar{\alpha}_t} x_0 + \sqrt{1 - \bar{\alpha}_t} \epsilon, \quad \epsilon \sim \mathcal{N}(0, I),
\end{equation}
where \( \bar{\alpha}_t \) is a predefined noise schedule. The model is trained to predict the noise \( \epsilon \) added at each timestep, thereby learning to reverse the noising process and sample clean data from pure noise.

\subsection{Label Diffusion for Generative Segmentation}

Unlike traditional diffusion models that operate over natural images, our formulation performs denoising in the \emph{label space}, where the clean signal \( x_0 \in \mathbb{R}^{\frac{H}{8} \times \frac{W}{8} \times 1} \) represents the ground-truth segmentation mask in latent form. At each diffusion timestep \( t \), we aim to recover \( x_0 \) from its noisy counterpart \( x_t \), conditioned on both the input image \( \mathcal{I} \) and the corresponding noun phrase \( T_i \). 

To this end, we model the conditional denoising process as:
\begin{equation}
\tilde{\epsilon}_t = \Phi\left( 
\text{concat}_\text{channel}(x_t,\ \varepsilon(\mathcal{I})),\ t,\ e^t_i,\ e^p_{il}
\right),
\label{eq:label-diffusion}
\end{equation}
where:
\begin{itemize}
    \item \( x_t \) is the noisy label map at timestep \( t \),
    \item \( \varepsilon(\mathcal{I}) \in \mathbb{R}^{\frac{H}{8} \times \frac{W}{8} \times 4} \) is the VAE-encoded image latent,
    \item \( \text{concat}(\cdot,\cdot) \) denotes channel-wise concatenation,
    \item \( e^t_i = \text{CLIP}(T_i) \) is the global text embedding of the noun phrase \( T_i \),
    \item \( e^p_{il} = \text{Adapter}_l(e^t_i) \) is the projected text feature injected into the U-Net at layer \( l \),
    \item \( \Phi \) is the denoising U-Net.
\end{itemize}

In this formulation, the image serves as a spatial condition via channel-wise concatenation to the noisy label, while the text provides semantic conditioning through cross-attention and multi-scale injection mechanisms. These two fusion pathways allow GS to align both spatial and linguistic cues when generating segmentation masks. Details of these fusion mechanisms are discussed in the next section.

\subsection{Multi-Scale Feature Injection}
As we mention, the text provides semantic conditioning through cross-attention and multi-scale injection mechanisms. Since the cross attention is traditional, we will only describe multi-scale injection here.

\paragraph{Latent Space Embedding.}
We adopt SDXL~\cite{rombach2022high} as the backbone diffusion model, operating in the latent space defined by a variational autoencoder \( \varepsilon \). The image \( \mathcal{I} \) is encoded to \( \varepsilon(\mathcal{I}) \in \mathbb{R}^{\frac{H}{8} \times \frac{W}{8} \times 4} \), which is concatenated with the noisy label map at each diffusion step.

\paragraph{Token-Level Adapter Injection.}
To enable fine-grained alignment between vision and language, we follow prior work~\cite{choi2024improving} and inject adapter features into the self-attention layers of the U-Net. Specifically, for each U-Net layer \( l \), the input feature map \( X_l \in \mathbb{R}^{N \times C} \) (flattened spatial tokens) is concatenated with the projected text features \( e^p_{il} \in \mathbb{R}^{M \times C} \) along the token dimension, yielding the joint token sequence:
\begin{equation}
A = \text{concat}_\text{token}(X_l,\, e^p_{il}) \in \mathbb{R}^{(N + M) \times C}.
\end{equation}
This joint sequence is then processed by the self-attention mechanism:
\begin{equation}
Z = \text{Attention}(A W^Q,\ A W^K,\ A W^V)\, W^O \in \mathbb{R}^{(N + M) \times C},
\end{equation}
where \( W^Q, W^K, W^V, W^O \) are the learned projection matrices. Finally, we discard the output tokens corresponding to the text embeddings and retain only the updated image features:
\begin{equation}
Y_l = Z_{1:N, :} \in \mathbb{R}^{N \times C},
\end{equation}
which are forwarded to the next U-Net layer. This design allows image tokens to attend to the injected language signals while preventing the injected tokens from contributing residual noise to downstream layers.

\subsection{Training and Inference with Classifier-Free Guidance}

\subsubsection{Training via Conditional Dropout.}
To enable classifier-free guidance~\cite{ho2022classifier}, we apply stochastic conditioning dropout during training. Specifically, with a fixed probability \( p_{\text{drop}} \), we replace the text embedding \( e^t_i \) and the projected adapter features \( e^p_{il} \) with learned null embeddings (``none" tokens), effectively simulating unconditioned denoising:
\vspace{-5pt}
\begin{equation}
    \tilde{\epsilon}_t = \Phi\left(
    \text{concat}_\text{channel}(x_t,\ \varepsilon(\mathcal{I})),\ t,\ \tilde{e}^t_i,\ \tilde{e}^p_{il}
    \right),
\end{equation}
where \( \tilde{e}^t_i \leftarrow \texttt{Drop}(e^t_i),\ \tilde{e}^p_{il} \leftarrow \texttt{Drop}(e^p_{il}) \), and \texttt{Drop} denotes random replacement with null tokens with probability \( p_{\text{drop}} \). The training objective minimizes the denoising prediction loss:

\begin{equation}
    \mathcal{L} = \mathbb{E}_{x_t, \epsilon} \left[ \| \epsilon - \tilde{\epsilon}_t \|^2 \right].
\end{equation}

\subsubsection{Inference via Classifier-Free Guidance.}
At inference time, we leverage both conditional and unconditional denoising predictions to improve semantic alignment. Specifically, we perform two forward passes of the U-Net at each timestep \( t \):
\vspace{-5pt}
\begin{align}
    \epsilon_{\text{cond}} &= \Phi(x_t,\ t,\ e^t_i,\ e^p_{il}), \\
    \epsilon_{\text{uncond}} &= \Phi(x_t,\ t,\ \texttt{None},\ \texttt{None}),
\end{align}
and apply the classifier-free guidance formula to interpolate:

\begin{equation}
    \epsilon_{\text{guided}} = \epsilon_{\text{uncond}} + w \cdot (\epsilon_{\text{cond}} - \epsilon_{\text{uncond}}),
\end{equation}
where \( w \geq 1 \) is the guidance scale. The denoising step then uses \( \epsilon_{\text{guided}} \) to update the latent:

\begin{equation}
    x_{t-1} = \texttt{DenoiseStep}(x_t,\ \epsilon_{\text{guided}},\ t),
\end{equation}
where \texttt{DenoiseStep} follows the standard DDPM update rule. After \( T \) steps, we obtain the final prediction \( \tilde{y}_i \) by decoding \( x_0 \) and resizing to the target resolution.

\begin{table*}[t]
\centering
\resizebox{0.99\width}{!}{
\begin{tabular}{c|c|c|c||ccccc}
\toprule[1.2pt]
\multirow{2}{*}{Method} & \multirow{2}{*}{Venue} & \multirow{2}{*}{Diffusion} & \multirow{2}{*}{P.S. Pretrain} & \multicolumn{5}{c}{Average Recall($\uparrow$)} \\ 
& & & & overall & things & stuff & singulars & plurals \\ \hline\hline
DiffSeg ~\cite{tian2024diffuse} & CVPR24 & \checkmark & $\times$ &  24.1 & 17.7 & 33.0 & 24.8 & 18.0 \\
DiffPNG ~\cite{yang2024exploring} & ECCV24 & \checkmark & $\times$ &  38.5 & 36.0 & 42.0 & 39.2 & 32.1 \\
EPNG~\cite{wang2023towards}& AAAI23  & $\times$ & $\times$ & 49.7 & 45.6 & 55.5 & 50.2 & 45.1 \\
MCN~\cite{luo2020multi} & CVPR20  & $\times$ & \checkmark & 54.2 & 48.6 & 61.4 & 56.6 & 38.8 \\
PNG~\cite{gonzalez2021panoptic} & ICCV21  & $\times$ & \checkmark & 55.4 & 56.2 & 54.3 & 56.2 & 48.8 \\
PPMN ~\cite{ding2022ppmn}  & ACMMM22   & $\times$ & $\times$ & 56.7 & 53.4 & 61.1 & 57.4 & 49.8 \\
EPNG ~\cite{wang2023towards} & AAAI23  & $\times$ & \checkmark & 58.0 & 54.8 & 62.4 & 58.6 & 52.1 \\
PPMN ~\cite{ding2022ppmn} & ACMMM22   & $\times$ & \checkmark & 59.4 & 57.2 & 62.5 & 60.0 & 54.0 \\
ODISE ~\cite{xu2023open} & CVPR23   & \checkmark & $\times$ & 61.0 & 57.0 & 66.6 & 61.7 & 54.8 \\
NICE ~\cite{wang2023nice} & arXiv & $\times$ & \checkmark & 62.3 & 60.2 & 65.3 & 63.1 & 55.2 \\
DRMN ~\cite{lin2023context} & ICDMI23 & $\times$ & \checkmark & 62.9 & 60.3 & 66.4 & 63.6 & 56.7 \\
ODISE ~\cite{xu2023open} & CVPR23 & \checkmark & \checkmark & 63.1 & 59.6 & 68.0 & 64.0 & 55.1 \\
PiGLET ~\cite{gonzalez2023piglet} & TPAMI23 & $\times$ & \checkmark & 65.9 & 64.0 & 68.6 & 67.2 & 54.5 \\
PPO-TD ~\cite{hui2023enriching} & IJCAI23 & $\times$ & \checkmark & 66.1 & 63.3 & 69.8 & 66.9 & 58.6 \\
EIPA+MLMA ~\cite{li2024dynamic} & ACMMM24 & $\times$ & \checkmark & 67.1 & 64.3 & 71.0 & 67.9 & 60.0 \\\hline\hline
GS(Ours) & - & \checkmark & $\times$ & \textbf{69.7} & \textbf{65.6} & \textbf{76.5} & \textbf{71.3} & \textbf{63.7} \\ 
  \bottomrule[0.5pt]
\end{tabular}
}
\caption{\textbf{Comparison with previous state-of-the-art methods on the PNG benchmark}. Our comparison on Average Recall metric includes 5 items: the overall average recall and four subcategories things and stuff categories, and singulars and plurals noun phrases.  ``P.S. Pretrain" denotes visual panoptic segmentation pretraining on COCO. The highest performances are reported among different model architectures.}
\label{tab:quantitative result_all}
\vspace{-11pt}
\end{table*}
\section{experiment}

\subsection{Datasets and Evaluation Protocol}
\subsubsection{Datasets} Following prior works, we train and evaluate our method on the \textbf{Panoptic Narrative Grounding} dataset. The PNG dataset establishes a large-scale multimodal benchmark by combining the narrative captions annotated in the \textbf{Localized Narratives} dataset with the panoptic segmentation annotated in the COCO dataset. The visual part of the dataset is based on the famous \textbf{COCO} dataset, with 133,103 images for training and 8533 images for validation. The PNG benchmark contains 726,445 noun phrases aligned with 741,697 segmentation masks, forming caption-image annotation pairs. Each caption averages 11.3 noun phrases, of which 5.1 necessitate grounding via segmentation masks. It should be noted that the noun phrases in the dataset are disaggregated into things and stuff categories, and singulars and plurals noun phrases.

\subsubsection{Metrics} The Average Recall (AR) metric in the PNG benchmark quantifies open-vocabulary segmentation recall robustness through multi-threshold IoU analysis. Computed as the mean recall across 10,000 distinct Intersection-over-Union thresholds ($\tau \in \{0.0001, 0.0002, \cdot, 0.9999\}$), AR evaluates a model’s ability to ground all noun phrases in narrative captions to precise spatial extents. For each noun phrase $T_{ip}$ and its ground-truth mask $y_p$, a predicted mask $\tilde{y}_p$ is considered correctly recalled at threshold $\tau$ iff $IoU(\tilde{y}_p, y_p) \ge \tau$. The final AR aggregates recall rates across all phrases and images:
\begin{align}
    AR=\frac{1}{10,000}\sum_{\tau}\frac{\sum_pCorrectRecallPhrases(\tau)}{\sum_pTotalPhrases(\tau)},
\end{align}
where $p$ presents the index of the picture.

Corresponding to the two paired attributes of noun phrases, the AR metric also comprises four sub-categories: $AR_{things}$, $AR_{stuff}$, $AR_{singulars}$ and $AR_{plurals}$. Each reflecting method performance on their respective categories.
\subsection{Implementation Details}
Our framework is implemented with PyTorch. We use SDXL inpainting model for the core diffusion pipeline and 2 different linear layers for the multi-scale adapter module. For stable diffusion, the total step is 1000, the DDIM diffusion step is 50 and the guidance scale is set to 7.5. Our model is trained using Adam optimizer with batch size of 128 and learning rate of 1e-7 for 10 epochs.

\subsection{Results and Analysis}
To evaluate our model, we conducted visual quality testing and quantitative metric testing on the PNG image dataset, achieving outstanding performance. Additionally, we tested it on real-world images, demonstrating the method's excellent generalization ability.

\begin{figure*}[t]
    \centering
    \includegraphics[width=1\linewidth]{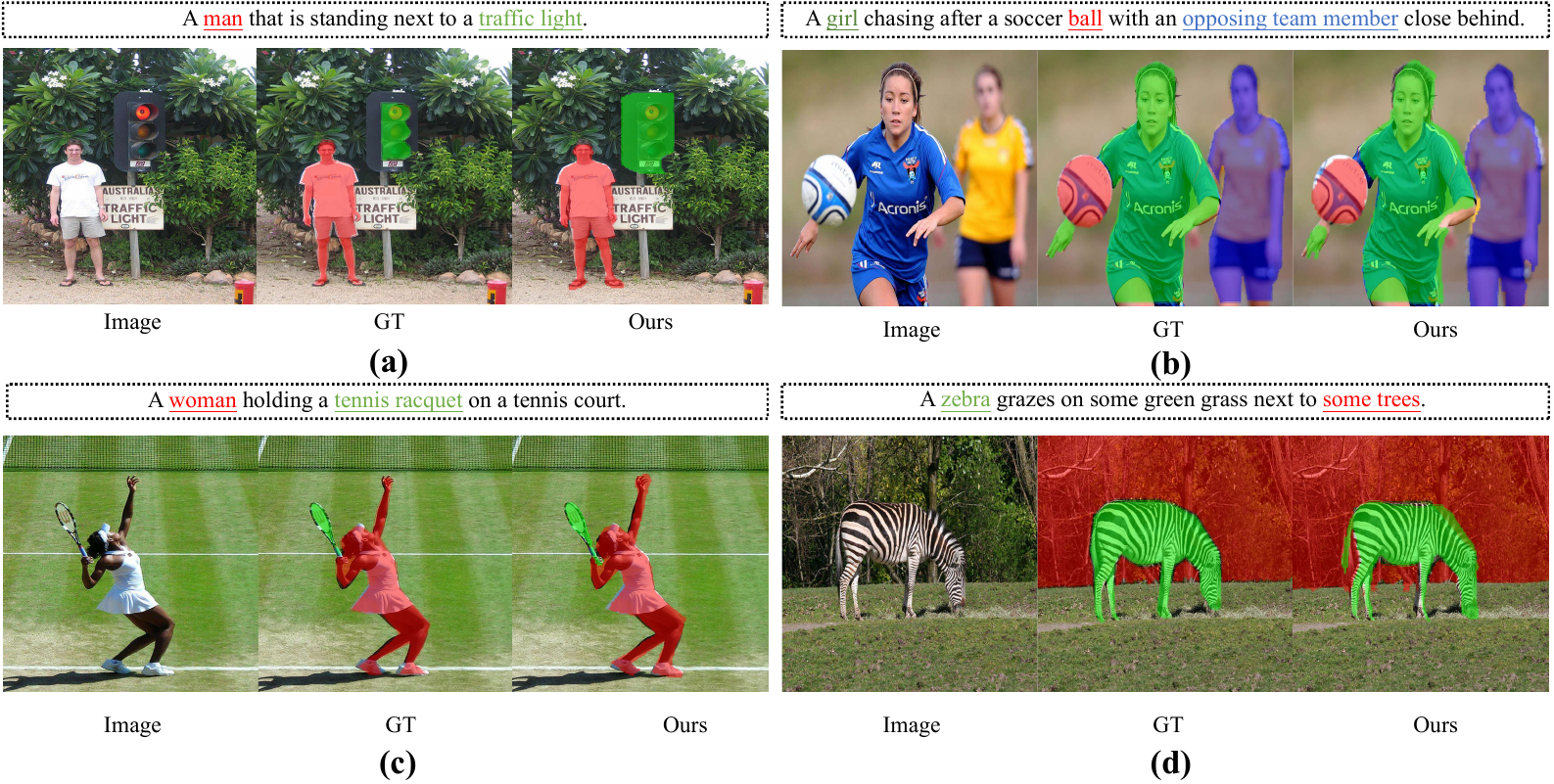}
    \caption{\textbf{Visualization of GS on representational samples on the PNG dataset.} All visualized images belong to the test set.}
    \label{fig:vis}
    \vspace{-5pt}
\end{figure*}
\begin{figure*}[t]
    \centering
    \includegraphics[width=1\linewidth]{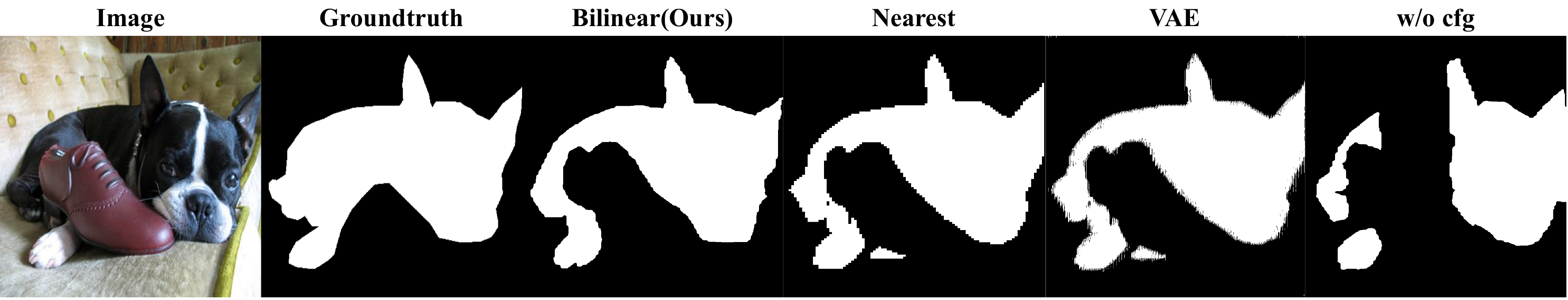}
    \caption{\textbf{Comparison with different post-processing techniques on final mask.} Notably, our bilinear interpolation method with classifier-free guidance (cfg) achieves superior results compared to gt. Furthermore, naive linear interpolation causes serrated edges, VAE decoding introduces undesired artifacts, and interpolation without cfg results in weaker conditional control.}
    \label{fig:mask_cmp}
    \vspace{-11pt}
\end{figure*}
\begin{figure*}[t]
    \centering
    \includegraphics[width=0.8\linewidth]{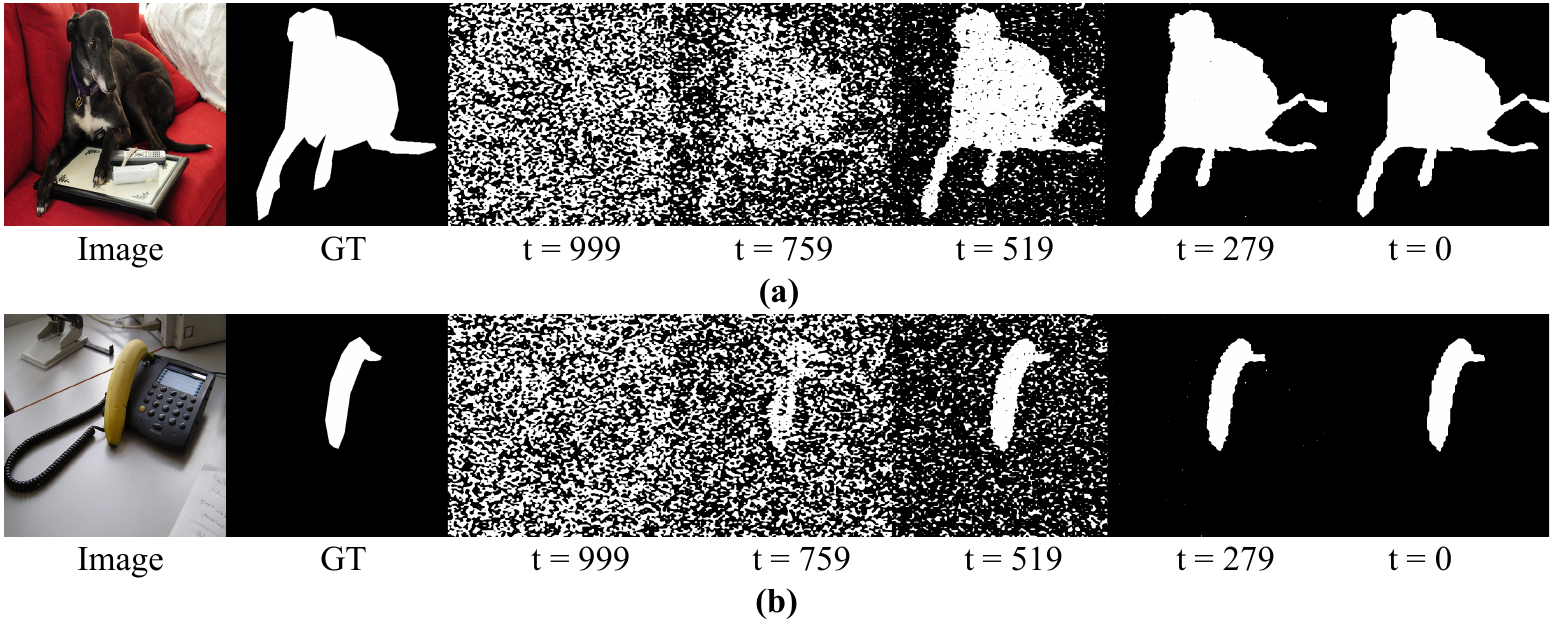}
    \vspace{-11pt}
    \caption{\textbf{Time-series Visualization of Images and Ground Truth for GS Samples on the PNG Dataset.} All visualized images belong to the test set. The figures are divided into two parts (a) and (b). Each part contains ``Image" and ``GT" columns, with the time steps t=0, t=279, t=519, t=759, and t=999 displayed to show the evolutionary process of the samples at different moments.}
    \label{fig:vis2}
    \vspace{-11pt}
\end{figure*}
\subsubsection{Qualitative Results.} In Fig. ~\ref{fig:vis}. we visualized the outputs of o
ur method on representative samples from the PNG dataset. Benefiting from the inherent smoothness of the diffusion model's generative process, our segmentation masks exhibit softer contours at the boundaries compared to the ground truth, thereby more precisely adhering to the target objects. For example, in (a), the image shows a man next to a traffic light and the ground truth failed to fully cover the target object, leaving an unlabeled perimeter around its edges, whereas our approach successfully segmented the entire traffic light. And in Fig. ~\ref{fig:mask_cmp}, the image depicts a dog lying on a sofa holding a shoe in its embrace. While the ground truth incorrectly segmented the upper portion of the shoe as part of the dog, our model remained unaffected by this ambiguity and robustly segmented the occluded shoe.

We then demonstrate the impact of different post-processing techniques on final mask quality in Fig. ~\ref{fig:mask_cmp}: First, using the VAE decoder to generate masks is a natural approach. However, as observed in our results, the VAE is susceptible to residual noise in the latent space, producing masks with artifacts. Second, nearest-neighbor interpolation results in more pronounced jagged edges compared to bilinear interpolation, degrading visual quality. Finally, inference without Classifier-Free Guidance (CFG) diminishes textual conditioning during generation, leading to detrimental visual outcomes.

\subsubsection{Quantitative Results.} As shown in table \ref{tab:quantitative result_all}, we presents a comparative analysis of various methods on the PNG benchmark using the Average Recall metric.

Among the evaluated methods, our proposed GS (Ours) achieves the state-of-the-art performance in terms of overall Average Recall, with a value of 69.7. When examining the sub - categories, it also outperforms other methods in ``stuff" (76.5), ``singulars" (71.3), and ``plurals" (63.7) Average Recall. The ``things" metric indicates our model underperforms counterparts on individual objects. Based on ablation studies across image sizes, we hypothesize that the constrained latent space dimensions in the VAE encoder impair this capability. Adopting larger image resolutions is expected to substantially enhance performance in this aspect.

As the first end-to-end generative approach, our GS achieves state-of-the-art performance among both diffusion-based and pixel-noun matching methods. Zero-shot Stable Diffusion frameworks like DiffSeg\cite{tian2024diffuse} and DiffPNG\cite{yang2024exploring} suffer from their non-trainable nature, yielding suboptimal results. Feature-extraction fine-tuning methods such as ODISE\cite{xu2023open} and EIPA+MLMA\cite{li2024dynamic} fall short of our approach. 

Notably, our method without pretraining outperforms other pretrained counterparts, even though the table shows that pretraining on COCO panoptic segmentation task significantly enhances model performance. This demonstrates the pioneering nature and effectiveness of our GS method in adapting diffusion models to the panoptic narrative grounding task.

\subsection{Ablation Study}
In this subsection, we conducted comprehensive ablation studies to investigate the impact of different design choices on segmentation quality. The ablation study was conducted on a subset of the test set, comprising approximately 7\% of its total size. The experiments focused on three key parameter factors: generated image size, DDIM steps, and guidance scale. 
\begin{table}[t]
\centering
\resizebox{0.99\columnwidth}{!}{
\begin{tabular}{c||ccccc}
\toprule[1.2pt]
\multirow{2}{*}{Size} & \multicolumn{5}{c}{Average Recall($\uparrow$)} \\ 
& overall & things & stuff & singulars & plurals \\ \hline\hline
(1024, 1024)  & \textbf{71.0} & \textbf{66.3} & \textbf{78.0} & \textbf{70.9} & \textbf{71.5} \\ 
(768, 768) &64.6 & 62.1 & 68.3 & 64.3 & 65.6\\
(512, 512) &32.7 & 24.4 &45.6 & 33.1 & 31.3 \\
  \bottomrule[0.5pt]
\end{tabular}
}
\caption{\textbf{Ablations on various image size}.}
\label{tab:ablation_imagesize}
\vspace{-11pt}
\end{table}
\begin{table}[t]
\centering
\resizebox{0.99\columnwidth}{!}{
\begin{tabular}{c||ccccc}
\toprule[1.2pt]
\multirow{2}{*}{Steps} & \multicolumn{5}{c}{Average Recall($\uparrow$)} \\ 
& overall & things & stuff & singulars & plurals \\ \hline\hline
50 & \textbf{71.0} & \textbf{66.3} & \textbf{78.0} & \textbf{70.9} & \textbf{71.5} \\
30  & 69.3 & 64.8 & 76.3 & 69.0 & 70.8 \\ 
20 & 68.7 & 63.9 & 76.2 & 68.5 & 69.9\\
  \bottomrule[0.5pt]
\end{tabular}
}
\caption{\textbf{Ablations on DDIM steps}}
\label{tab:ablation_ddimSteps}
\vspace{-11pt}
\end{table}
\begin{table}[t]
\centering
\resizebox{0.99\columnwidth}{!}{
\begin{tabular}{c||ccccc}
\toprule[1.2pt]
\multirow{2}{*}{Guidance Scale} & \multicolumn{5}{c}{Average Recall($\uparrow$)} \\ 
& overall & things & stuff & singulars & plurals \\ \hline\hline
7.5  & \textbf{71.0} & 66.3 & \textbf{78.0} & \textbf{70.9} & \textbf{71.5} \\ 
8.0  & 70.5 & 66.9 &76.3 & 70.4 & 71.3\\
7.0 & 70.7 & \textbf{67.0} & 76.6 & 70.7 & 71.1\\
  \bottomrule[0.5pt]
\end{tabular}
}
\caption{\textbf{Ablations on Guidance Scale.}}
\label{tab:ablation_guidancescale}
\vspace{-11pt}
\end{table}
\subsubsection{Image Size.} This subsection investigates the critical impact of the generated image size on the performance and output quality of GS. In table ~\ref{tab:ablation_imagesize}, we can see the overall accuracy decreases proportionally with the reduction in generated image size. We attribute this performance degradation to insufficient spatial information in undersized images. Due to the limited receptive field, the model produces less precise masks, which particularly impacts the 'things' and 'singulars' subcategories that require higher segmentation precision.

\subsubsection{DDIM Steps.} Here, we focus on ablating the number of sampling steps in the denoising diffusion implicit model (DDIM) process. In table ~\ref{tab:ablation_ddimSteps}, the increasing the number of sampling steps allows for finer-grained capture of the data distribution, potentially enhancing both the accuracy and detail of the generated results. By incrementally reducing the step count from a baseline high value, we observe a slight performance degradation in the model, which aligns with theoretical expectations of DDIM.

\subsubsection{Guidance Scale.} The third ablation study examines the influence of the classifier-free guidance scale, a key hyperparameter controlling the trade-off between sample diversity and fidelity to the conditioning signal (e.g., text prompt). We conducted exploratory investigations in the neighborhood of the empirically determined guidance scale parameter, which confirmed the optimality of the original empirical value.

\subsection{Appliable Efficiency.}
Our accurate segmentation approach consists of a lightweight adapter and a performance-optimized SDXL pipeline. On an A100 GPU, a single inference pass with a batch size of 16 takes only 50 seconds, averaging just 3 seconds per image generated.

\section{Conclusion}
\label{sec:conclusion}

We introduced \textbf{GS}, a novel framework that formulates language-driven segmentation as a generative task via \emph{label diffusion}. In contrast to existing image-centric approaches that treat segmentation as a downstream or auxiliary objective, GS directly generates segmentation masks from noise conditioned on both visual and linguistic inputs. This reorientation makes the label space the central modeling target, enabling more faithful spatial and semantic grounding.

To support this paradigm, we designed a dual conditioning mechanism that integrates image structure and textual semantics into the generative process through latent fusion and multi-scale adapter injection. Experimental results on the Panoptic Narrative Grounding benchmark demonstrate that GS achieves new state-of-the-art performance, highlighting the effectiveness and generality of treating segmentation as generative modeling. We believe GS opens a promising direction for unifying generative modeling and structured prediction in multimodal tasks.

\bibliography{aaai2026}

\clearpage
\appendix
\section{Supplementary Content}

This is the supplementary material of paper \textbf{GS: Generative Segmentation via Label Diffusion}. We have included additional favorable visualization samples from the test set, along with problematic cases for reviewers' analysis. The source code and deployment instructions are also provided.

\begin{figure*}[t]
    \vspace{-20pt}
    \centering
    
    \includegraphics[width=0.9\linewidth]{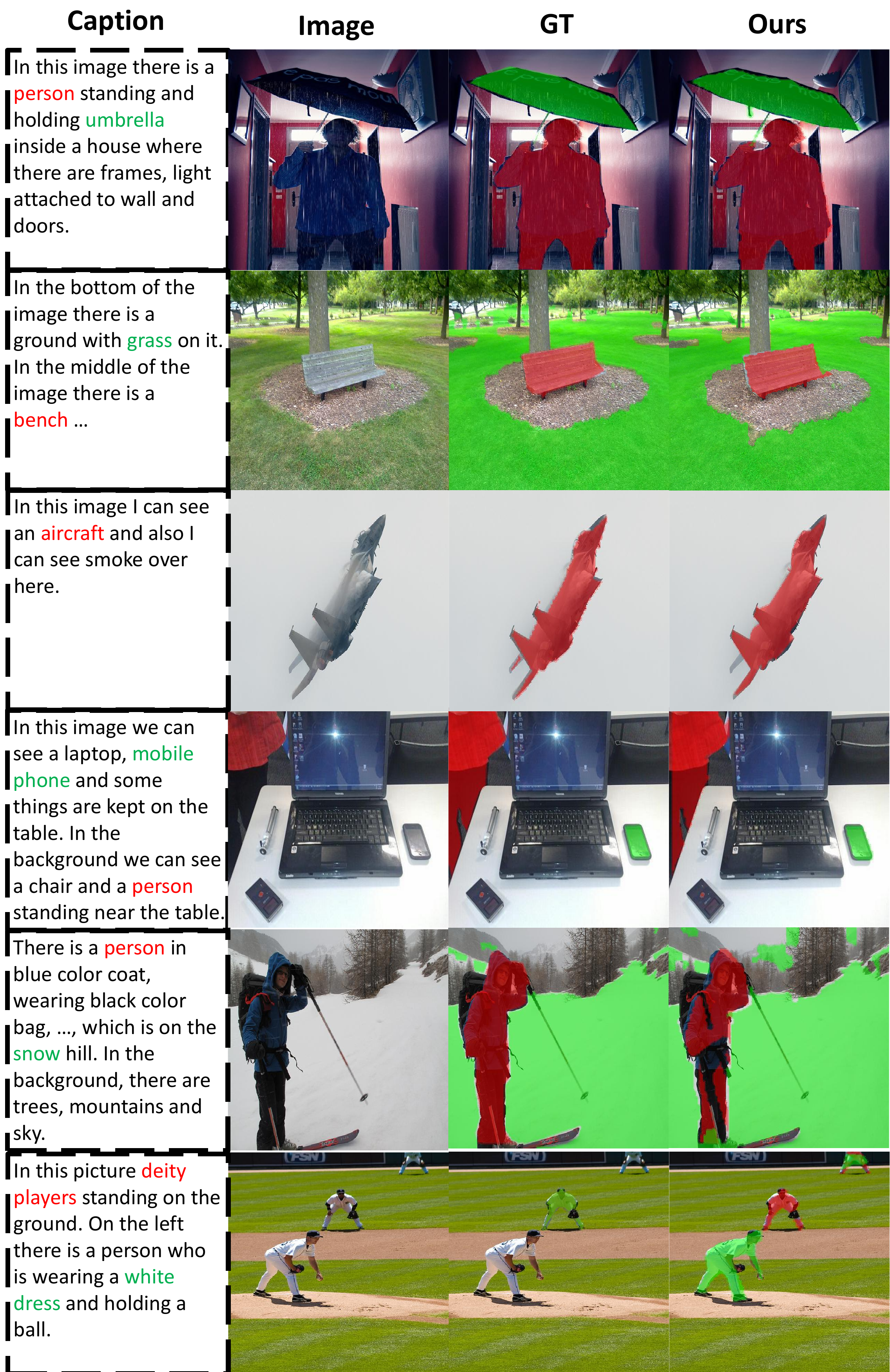}
    \caption{\textbf{Visualization of Images and Ground Truth for Correct GS Samples on the PNG Dataset test split.} }
    
    \label{fig:Correct}
    \vspace{-20pt}
\end{figure*}
\section{Correct samples}
In Fig. ~\ref{fig:Correct}, we provide more samples that our model generates. In the \textbf{First Line} example, the two target prompts ``person" and ``umbrella" exhibit similar coloration and are spatially adjacent. Nevertheless, our model successfully delineates both targets with high precision through robust semantic alignment, effectively resisting visual interference. In the  \textbf{Fourth Line} example, one target prompt ``person" is neither centrally positioned nor constitutes a complete human figure; the other target ``mobile phone" occupies minimal image area. Our model successfully segments both targets with high fidelity, validating its generalization capability for peripheral and partial objects, along with precise segmentation of small-scale objects.
\begin{figure*}[t]
    \centering
    \includegraphics[width=0.9\linewidth]{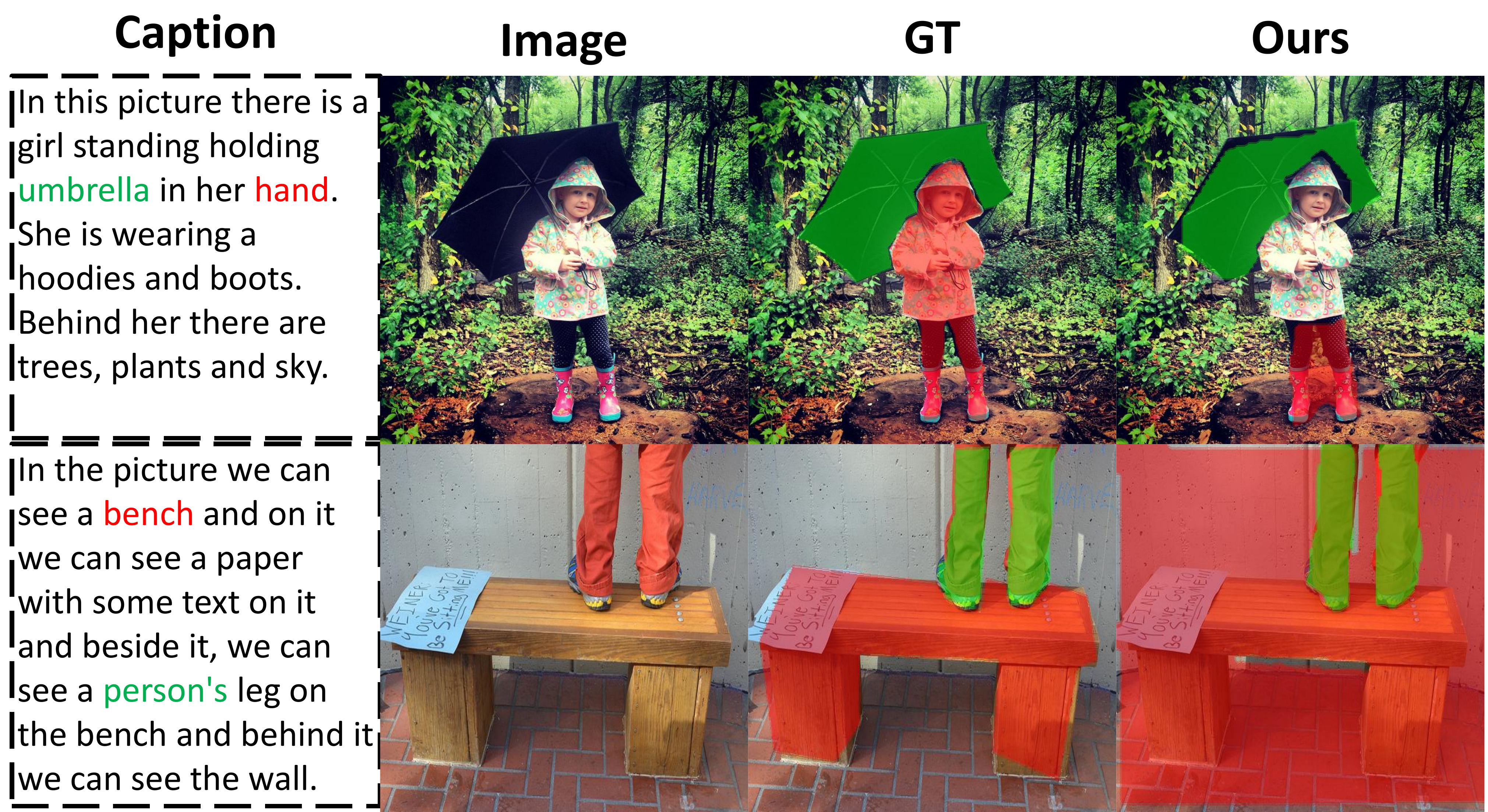}
    \caption{\textbf{Visualization of Images and Ground Truth for Failed GS Samples on the PNG Dataset test split.} }
    \label{fig:failure}
\end{figure*}
\section{Failed samples}
In Fig. ~\ref{fig:failure}, we provide some samples that our model failed to predict correct mask. In the \textbf{UP} example, for the segmentation target ``hand", the ground truth captures the entire girl, whereas our method erroneously segments only the lower body of the child. In the \textbf{DOWN} example, for the segmentation target ``bench", the ground truth correctly delineates the entire bench, whereas our method catastrophically classifies almost the entire scene as the bench.

\end{document}